# VLM-MPC: Model Predictive Controller Augmented Vision Language Model for Autonomous Driving

Keke Long[1], Haotian Shi[1], Jiaxi Liu[1], Chaowei Xiao[1], Xiaopeng Li[1]*

[1] Department of Civil and Environmental Engineering, University of Wisconsin-Madison, Madison, WI, 53706, USA

Corresponding author: xli2485@wisc.edu

**ABSTRACT**

Motivated by the emergent reasoning capabilities of Vision Language Models (VLMs) and their potential to improve the comprehensibility of autonomous driving systems, this paper introduces a closed-loop autonomous driving controller called VLM-MPC, which combines the Model Predictive Controller (MPC) with VLM to evaluate how model-based control could enhance VLM decision-making. The proposed VLM-MPC is structured into two asynchronous components: The upper layer VLM generates driving parameters (e.g., desired speed, desired headway) for lower-level control based on front camera images, ego vehicle state, traffic environment conditions, and reference memory; The lower-level MPC controls the vehicle in real-time using these parameters, considering engine lag and providing state feedback to the entire system. Experiments based on the nuScenes dataset validated the effectiveness of the proposed VLM-MPC across various environments (e.g., night, rain, and intersections). The results demonstrate that the VLM-MPC consistently maintains Post Encroachment Time (PET) above safe thresholds, in contrast to some scenarios where the VLM-based control posed collision risks. Additionally, the VLM-MPC enhances smoothness compared to the real-world trajectories and VLM-based control. By comparing behaviors under different environmental settings, we highlight the VLM-MPC's capability to understand the environment and make reasoned inferences. Moreover, we validate the contributions of two key components, the reference memory and the environment encoder, to the stability of responses through ablation tests.

**Keywords:** Autonomous Vehicle, Foundation Model, Large Language Model, Vehicle Safety

# 1 INTRODUCTION

The comprehensibility of autonomous driving systems is of paramount importance. Existing autonomous driving systems, whether model-based (Gao et al., 2022; Shang et al., 2022) or learning-based (Shi et al., 2021), typically require complex rules or reward function designs to understand the traffic environment and driving preferences. However, this reliance on predefined rule sets often limits their adaptability to various traffic scenarios (Tian et al., 2024). Furthermore, learning-based autonomous driving systems face the challenge of out-of-distribution (OOD) data (Filos et al., 2020), where limited datasets fail to ensure reliable decision-making in rare real-world driving situations.

Given these challenges, foundation models (FMs) show potential as a promising solution (Brown et al., 2020; Valmeekam et al., 2023). Leveraging extensive datasets of text and images, FMs have endowed them with reasoning capabilities (Brown et al., 2020; Ma et al., 2023), enhancing the interpretability and performance of autonomous driving systems in complex environments. Recent studies have begun utilizing Vision Language Models (VLMs) and Large Language Models (LLMs) to process driving environmental inputs, such as images and the positions of traffic participants. This information is encoded as language representations and input into FMs, enabling them to perform driving tasks effectively.

The application of FMs in autonomous driving tasks encompasses various roles. Depending on the control mode output by the FM, current FM-based vehicle controllers can be classified into two categories, as outlined in TABLE 1. The first category of approach, "FM to action" (Sha et al., 2023), directly uses FMs to generate autonomous vehicle (AV) actions, such as waypoints, throttle, and pedal commands. This is an 'end-to-end' autonomous driving mode. However, these studies often fail to account for vehicle kinematics and dynamics in FM models (Mao et al., 2023), raising concerns about the real-world applicability of the generated actions. Additionally, this approach poses challenges to the response speed of FMs. In autonomous driving, particularly in near-collision scenarios, a control frequency of approximately 10Hz is required (Campolo et al., 2017). However, current FMs are incapable of achieving such high response speeds, further limiting their practicality in real-world applications. Some studies have sacrificed model accuracy by selecting smaller models for faster response times.

To leverage FMs' reasoning capabilities while mitigating the issue of slow response speeds, some AV control studies have adopted the second category of approach: "FM to commands/codes." In this approach, FMs serve as upper-level decision-makers, providing instructions or commands to the lower control system. This method decouples upper-level FM from lower-level vehicle control, allowing the upper-level FM to focus on long-term planning at a slower frequency. Meanwhile, the lower level inherits mature control schemes to perform high-frequency control. This asynchronous control approach offers two key benefits. First, it reduces the necessity for high response speeds from the FM, making it more compatible with the current capabilities of FMs (Chen et al., 2024). Second, it allows the lower-level control to leverage well-established vehicle control schemes, such as model predictive control (MPC)-based trajectory tracking (Wang et al., 2021), which are widely adopted in the industry.

However, a fundamental question remains unanswered: does incorporating a model-based approach like MPC into an FM benefit the FM's performance in autonomous driving? To tackle this unsolved question, we propose an autonomous driving controller with a two-level structure: VLMs, representing FM, serve as the upper-level vehicle control, while MPC functions as the lower-level controller. The upper-level VLM processes inputs such as the vehicle's front camera images, textual environment descriptions, and reference memory to generate control parameters for the lower-level MPC. The MPC then uses these parameters, accounting for vehicle dynamics with engine lag, to achieve realistic vehicle behavior and provide state feedback to the VLM. This asynchronous two-layer structure effectively mitigates the challenge of slow VLM response speeds. We compare this controller with the VLM-to-action approach to evaluate the impact of integrating MPC within a VLM framework.

TABLE 1 Related Research

| | Research | Foundation model Input | Foundation model Output | Open/ Closed loop | Data/ Platform |
|---|---|---|---|---|---|
| FM to action | OmniDrive (S. Wang et al., 2024) | 3d vision/traffic info/trajectory memory/instruction | Planned waypoint | Open | nuScenes |
| | 3D-Tokenized LLM (Bai et al., 2024) | 3d vision/traffic info/trajectory memory/instruction | Planned waypoint | Open | nuScenes |
| | GPT-Driver (Mao et al., 2023) | Text scene description (status of ego vehicle and surrounding vehicle) | Description of the action; trajectory | Open | nuScenes |
| | Language Agent (Mao et al., 2024) | Text scene description (status of ego vehicle and surrounding vehicle) | Description of the action; trajectory | Open | nuScenes |
| | DriveGPT (Xu et al., 2024) | Videos, text instructions | Control (speed, turning angle) | Open | CC3M; WebVid-2M; BDD-X |
| | BEVGPT (P. Wang et al., 2024) | Bird's-eye-view (BEV) images | Trajectory | Open | Lyft |
| | Driving with LLMs (Chen et al., 2023) | Text scene description (route, vehicle, pedestrian, ego state) | Steering/pedal | Open | Self-made open dataset |
| | SurrealDriver (Jin et al., 2023) | Text memory, safety criteria, text scenario description | Action (speed up/stop) | Close | Carla |
| | ADriver-I (Jia et al., 2023) | Historical frames, text scenario description | speed, Steer Angle | Close | nuScenes |
| | LMDrive (Shao et al., 2023) | Traffic info, vision, instruction | Waypoints | Close | Carla |
| | DriveVLM (Tian et al., 2024) | Traffic info/vision/ instruction | Waypoints; reference trajectory for tracking | Close | nuScenes, SUP-AD dataset |
| FM to comm- ands/c odes | DiLu (Wen et al., 2023) | Text Text scene description | Five operational actions | Open | Highway-env, CitySim |
| | Personalized autonomous driving (Cui et al., 2024b) | Text scene description, instructions, API docs | Language model program | Open | Self-made open dataset |
| | languageMPC (Sha et al., 2023) | Text scene description | MPC parameters | Open | IdSim |
| | LLM-Assist (Sharan et al., 2023) | Driving parameters | Trajectory planner parameter | Close | nuPlan |
| | PlanAgent (Zheng et al., 2024) | BEV map, text scene description | Planning parameters | Close | nuPlan |
| | Co-driver (Guo et al., 2024) | Front camera image, text scene description | Control type, control parameter | Close | Carla |

| Asynchronous llm (Chen et al., 2024) | Feature for trajectory planner | Driving parameters | Close | nuPlan |
| DriVLMe (Huang et al., 2024) | Perception video, text history action, planned route, human instruction | Eight operational actions | Close | Carla |
| Drive as You Say (Cui et al., 2024a) | Text scene description | Five operational actions | Close | HighwayEnv |
| LaMPilot (Ma et al., n.d.) | Text scene description, instructions, API docs | Code | Close | Self-made open dataset |

Another critical problem for FMs is the occurrence of hallucinations in their outputs (Farquhar et al., 2024). Hallucinations are often defined as LLMs generating 'nonsensical or unfaithful content to the provided source content' (Ji et al., 2023). These can arise for various reasons, including biases in the training data, lack of context awareness, and inherent limitations in the model's architecture. These hallucinations hinder the application of FMs across various fields, including autonomous driving, where they may fail to accurately comprehend the driving environment, leading to inconsistent responses or the generation of misleading information. Currently, a systematic evaluation to verify solutions that could enhance the reliability of FM responses in autonomous driving tasks is relatively rare.

In summary, current studies in FM research for autonomous driving control have not discussed the contribution of model-based controllers like MPC to FM-based control and lack a discussion about the methods for improving the reliability of FM responses in autonomous driving. To fill this gap, we propose a VLM-MPC framework for autonomous driving. The proposed VLM-MPC is structured into two asynchronous layers: The upper layer VLM generates driving parameters for lower-level control, and The lower-level MPC controls the vehicle in real-time using these parameters.

To evaluate the performance of VLM-MPC against the FM-to-action approach, a real-world driving scenarios dataset is adopted for testing. The results demonstrate that VLM-MPC ensures safety and improves smoothness across all tested scenarios, whereas the baseline FM-to-action method poses safety risks. To further validate the proposed VLM-MPC, we extend the analysis to a different variation: the LLM-MPC framework, which also demonstrates safety and improved smoothness across tested scenarios. Furthermore, we emphasize environmental understanding by VLM by examining its performance in different scenarios and the visualization of VLM attention to the image input. Results show that the VLM exhibited appropriate variations in behavior under different environmental conditions.

To avoid hallucination and improve the response stability of VLM, the VLM-MPC framework incorporates two key components: the reference memory and the environment encoder. The reference memory compiles average parameters applicable to current scenes based on their characteristics. The environment encoder extracts vital environmental factors such as weather, lighting, and road conditions from images captured by the vehicle's front camera. These two components provide additional context to the VLM, thus mitigating the risk of incorrect or nonsensical responses. Ablation tests validate that both components significantly enhance the response stability of VLM, thereby improving the overall performance and reliability of VLM-MPC.

The structure of this paper is as follows. Section 2 describes the proposed VLM-MPC method. Section 3 shows the experiment settings and results. Section 4 concludes the paper and discusses the limitations of the proposed method and future research. Our code and data will be made available upon acceptance of this paper.

## 2 METHODOLOGY

This section presents VLM-MPC, a VLM-based controller for autonomous driving. We first introduce the overall architecture of VLM-MPC in Section 2.1. Then, we introduce the two levels of our method: upper-layer VLM (Section 2.2) and lower-layer MPC controller (Section 2.3)

### 2.1 Overall Architecture

The overall pipeline of the proposed closed-loop VLM-MPC is shown in Figure 1. The VLM-MPC is structurally divided into two levels: an upper-level VLM part serves as an upper-level planner, and a lower-level MPC part for autonomous vehicle control.

**Upper-level VLM**: The upper level relies on VLM as the core reasoning and decision-making component, consisting of four subcomponents: 1) Environment encoder (Section 2.2.1): Generates environment descriptions from the vehicle's front camera. 2) Scene Encoder (Section 2.2.2): Extracts information about the ego vehicle and preceding vehicles from the dataset. 3) Reference memory (Section 2.2.3): This component stores aggregated driving parameters from the history driving dataset, providing the VLM with historical context and average parameters for varied driving scenarios. 4) Prompt Generator: Constructs prompts for input and guides the VLM to perform chain-of-thought (CoT) analysis. The first three components collect information to construct the prompt for the fourth component. Once the prompt is input into the VLM, it produces the upper-level results, specifically the key driving parameters.

**Lower level MPC**: The lower level is a rule-based MPC. This part considers the vehicle's underlying dynamics. It relies on the key driving parameters set by the upper level to determine specific actions, which are then transmitted to the vehicle for execution (Section 2.3.1).

The upper and lower levels of the VLM-MPC are decoupled and operate at different frequencies. This setup enables controlled asynchronous inference. Thus, the upper-level VLM is not required to process at a high frequency as the lower-level MPC. During asynchronous intervals, the previously derived upper-level instruction and driving parameters continue to guide the lower-level MPC.

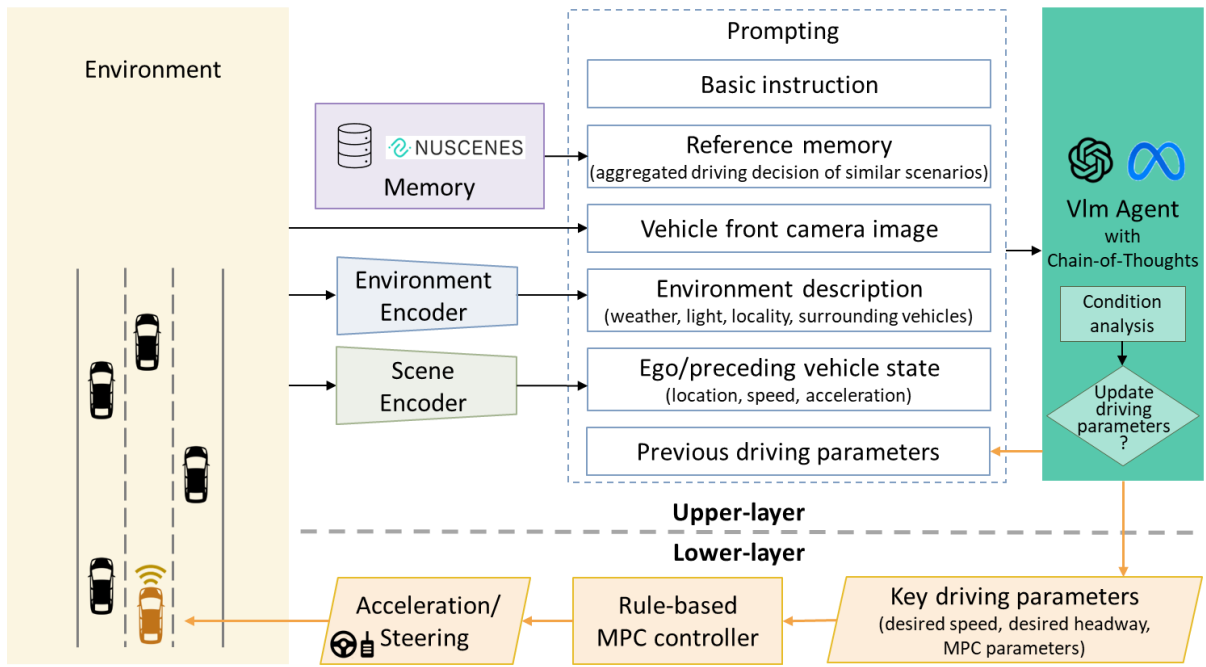

Figure 1 Architecture of VLM-MPC.

## 2.2 Upper-layer: VLM

This section introduces how the upper-layer: VLM, collects scene information and constructs prompts for the VLM to generate key driving parameters. This section introduces each component in sequence: reference memory dataset, environment encoder, scene encoder, and prompt generator.

### 2.2.1 Environment Encoder

Driving environment characteristics, such as weather, light, and road conditions, significantly influence driving performance. Let $\mathcal{T} \coloneqq \{t_1, t_2, \ldots, t_n\}$, where n represents the total number of time points required to complete a driving task within one scene. To enhance the environmental understanding of VLM, this model outputs a linguistic description $E_t$ of the driving environment at time $t \in \mathcal{T}$. $E_t$ includes 6 detailed scene descriptions at time $t$:

$$E_t \coloneqq \{E_t^{Wea}, E_t^{Lig}, E_t^{RT}, E_t^{RC}, E_t^{Obs}\}, t \in \mathcal{T} \tag{1}$$

The Contrastive Language-Image Pre-training (CLIP) model (Fang et al., 2024; Radford et al., 2021) is adopted to extract $E_t$ from the vehicle front camera imagesA pre-trained 'ViT-B/32' configuration is loaded. Several categories of descriptive texts were defined, including weather conditions, lighting, road types, road conditions, and static obstacles. The process begins by preprocessing and encoding an input image using the CLIP model to obtain image features. These features are then compared with pre-defined descriptive texts through cosine similarity calculations between the image and text features encoded by the CLIP model, as shown in Figure 2.

In this setup, descriptions of weather conditions $E_t^{Wea}$, lighting $E_t^{Lig}$, and road types $E_t^{RT}$ are outputted regardless of the confidence score, ensuring these essential scene attributes are consistently provided. For road conditions $E_t^{RC}$ and obstacles $E_t^{Obs}$, the descriptions are only outputted if their confidence scores exceed specific thresholds (0.3 and 0.2, respectively). This conditional output prevents the inclusion of less specific information, maintaining the accuracy and relevance of the extracted features.

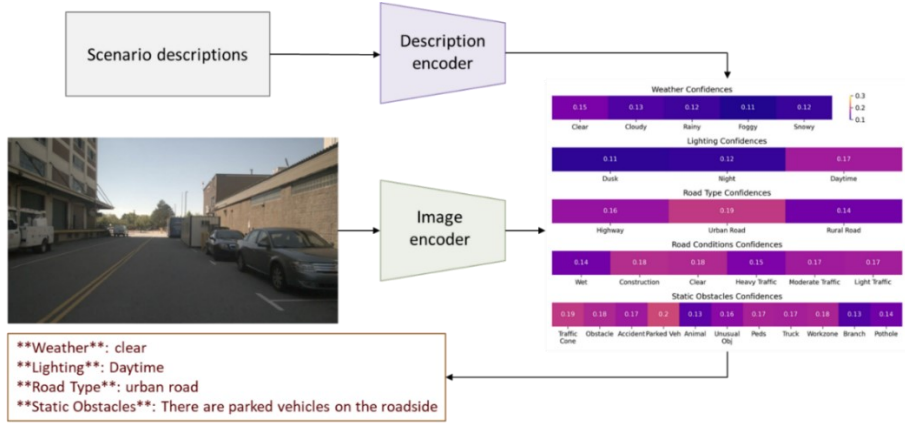

Figure 2 CLIP model structure in environment encoder

### 2.2.2 Scene Encoder

To enhance the VLM's understanding of the driving task, the prompt includes the historical state information of the ego vehicle and the states of significant surrounding vehicles and other obstacles. Significant surrounding vehicles refer to the preceding vehicle that the MPC algorithm needs to consider for safe and efficient car-following behavior. Other obstacles include any objects that obstruct the ego vehicle's path, such as pedestrians, and elements that need to be considered due to traffic rules, such as stop lines at red lights or stop signs at intersections.

The scene encoder extracts the scene status $S_t$ at time $t$, including the ego vehicle state $s_t^{\text{ego}}$, the preceding vehicle state $s_t^{\text{pre}}$ (if applicable), and the position of the stop line $s_t^{\text{SL}}$ (if applicable):

$$S_t \coloneqq \{s_t^{\text{ego}}, s_t^{\text{pre}}, s_t^{\text{SL}}\}, t \in \mathcal{T} \tag{2}$$

$$s_t^{\text{ego}} \coloneqq \left[x_{t' \in \mathcal{T}_t}^{\text{ego}}, v_{t' \in \mathcal{T}_t}^{\text{ego}}, a_{t' \in \mathcal{T}_t}^{\text{ego}}\right]_{\mathcal{T}_t = \{t-\gamma\Delta t, \cdots, t-\Delta t, t\}}, t' \in \mathcal{T} \tag{3}$$

$$s_t^{\text{pre}} \coloneqq \left[\Delta x_t^{\text{pre}}, v_t^{\text{pre}}, a_t^{\text{pre}}\right], t \in \mathcal{T} \tag{4}$$

$$s_t^{\text{SL}} \coloneqq \left[\Delta x_t^{\text{SL}}\right], t \in \mathcal{T} \tag{5}$$

where $s_t^{\text{ego}}$ are the $\gamma$ steps of historical states of the ego vehicle containing vehicle position, vehicle speed, and vehicle acceleration at the previous time set $\mathcal{T}_t = \{t - \gamma\Delta t, \cdots, t - \Delta t, t\}$, $\Delta t$ is the historical state sample interval. In this research, $\gamma = 5$, $\Delta t = 0.5s$. $s_t^{\text{pre}}$ is the status information of the preceding vehicle at time $t$. $\Delta x_t^{\text{SL}}$ is the vehicle position in Frenet coordinates (Reiter and Diehl, 2021), with the ego vehicle as the origin. $s_t^{\text{SL}}$ is the status information of the stop line, $s_t^{\text{SL}}$ exists when the lane of the ego vehicle is blocked by a traffic signal or stop sign.

The preceding vehicle's information is determined based on vehicle positions, road information, and other vehicle positions from the nuScenes dataset (Caesar et al., 2020). Similarly, stop line positions are derived from the map information provided in the nuScenes dataset. This study does not account for errors in the perception phase of traffic participants.

### 2.2.3 Reference Memory

Reference memory is constructed based on real vehicle trajectories from the nuScenes dataset. It serves as a reference for generating the key driving parameters by aggregating data from various driving scenes. Although the trajectories from nuScenes are not exclusively from autonomous driving or human driving, they generally exhibit safe driving behaviors, as supported by the safety analysis in the experiment section. Thus, we consider these trajectories valuable references for determining key driving parameters.

The key driving parameters include six parameters: $[N, Q, R, Q^h, v^d, h^d]$:

1. **Prediction horizon** $N$: The number of time steps the controller looks ahead.
2. **Speed maintenance weight** $Q$: Weight for maintaining the desired speed. Considering there are three weights in the system, $Q$ is fixed at 1.
3. **Control effort weight** $R$: Weight for minimizing control effort.
4. **Headway maintenance weight** $Q^h$: Weight for maintaining the desired headway (distance) to the front vehicle.
5. **Desired speed** $v^d$: The target speed (m/s) for the ego vehicle.
6. **Desired headway** $h^d$: The desired headway (s) between the ego vehicle and the front vehicle.

The six parameters are selected to optimize autonomous vehicle control. The prediction horizon $N$ anticipates future changes, enabling proactive adjustments. Speed maintenance weight $Q$ and control effort weight $R$ balance desired speed with minimal control effort for efficient driving. Headway maintenance weight $Q^h$ ensures a safe distance from the vehicle ahead, which is crucial for collision avoidance and traffic flow. Desired speed $v^d$ and desired headway $h^d$ set specific performance targets. Together, these parameters harmonize safety, efficiency, and comfort, enhancing the driving experience.

The reference memory dataset encapsulates these parameters using the notation $M(E_t)$, the function $M$ maps each scenario to a tuple of averaged MPC parameters specific to that scenario:

$$M(E_t) = \left[\bar{N}(E_t), \bar{Q}(E_t), \bar{R}(E_t), \bar{Q}^h(E_t), \bar{v}^d(E_t), \bar{h}^d(E_t)\right], t \in \mathcal{T}$$

Here $\bar{N}(E_t)$ denote the average values of the prediction horizon for all scenarios that share the same environment characteristics of $E_t$, Similarly, $\bar{Q}(E_t), \bar{R}(E_t), \bar{Q}^h(E_t), \bar{v}^d(E_t)$ and $\bar{h}^d(E_t)$ represent the average values of the speed maintenance weight, control effort weight, headway maintenance weight, desired speed, and desired headway, respectively, for those scenarios.

To build the reference memory dataset, we calibrated the MPC parameters for each scene in the nuScenes dataset by extracting specific characteristics such as intersection, rain, and night. Each scene was categorized based on a combination of these characteristics, resulting in eight distinct scenarios (e.g., rain at an intersection at night). We then performed statistical analyses to determine whether these scenario pairs had significant differences in driving behaviors. If significant differences were found, different average driving parameters were adopted for each scenario; if not, the scenarios were combined, and the same average driving parameters were used. This approach ensures that the reference memory provides appropriately tailored parameters for various driving conditions.

### 2.2.4 Prompt Generator

The prompt generator collects all relevant information from the various modules and generates the input prompts for the VLM. This process ensures the VLM can effectively generate the six necessary parameters at time $t$:

$$\theta_t = [N, Q, R, Q^h, v^d, h^d] \tag{6}$$

To generate these parameters, there is a slight difference in how the prompt is constructed during the initial call to the VLM for each scene and subsequent updates. For each scene, the first call at time $t = 0$ to the VLM for upper-level decision-making includes the following inputs: memory $M$, the front camera frame $img_{t=0}$, environment description $E_{t=0}$, and the state of the ego vehicle and preceding vehicles $s_{t=0}$. The initial driving parameters $\theta_{t=0}$ are calculated as:

$$\theta_{t=0} = f^{VLM}(M(E_{t=0}), img_{t=0}, E_{t=0}, S_{t=0}) \tag{7}$$

Subsequent calls to the VLM at intervals of $\Delta t^u$ for updating decisions, including the front camera frame $img_t$, environment description $E_t$, the state of the ego vehicle and preceding vehicles $s_t$, and the previous driving parameters $\theta_{t-\Delta t^u}$. The updated driving parameters $\theta_t$ are calculated as:

$$\theta_t = f^{VLM}(img_t, E_t, S_t, \theta_{t-\Delta t}), t \in \{\Delta t^u, 2\Delta t^u, \cdots\} \tag{8}$$

At the end of the prompt, we utilize chain-of-thought (CoT) prompting—presenting a series of logical and interconnected steps or directives—to help the VLM better navigate the complexities of real-world driving tasks and conduct human-like reasoning. The CoT guidance is shown in Figure 3. This method enhances the VLM's ability to handle complex scenarios by breaking the decision-making process into manageable, logical steps, improving overall performance in dynamic environments.

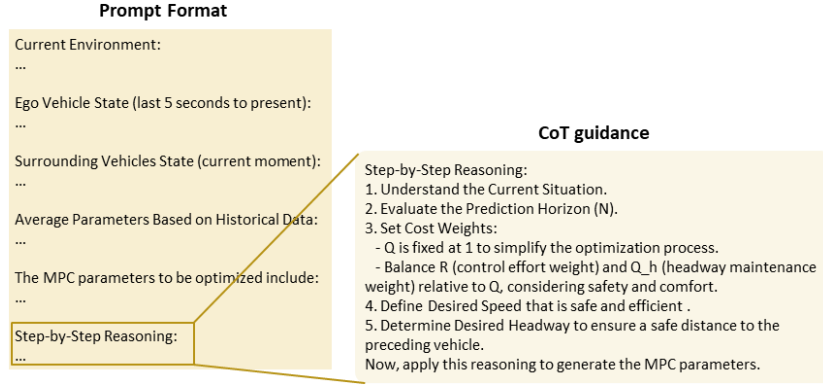

Figure 3 General prompt structure and the CoT guidance in prompts.

## 2.3 Lower-layer: MPC

In the lower layer, we account for the vehicle dynamics, including engine lag. Building on the key parameters $\theta_t$ generated by the VLM, the MPC component optimizes the trajectory of the controlled vehicle in real time. The MPC model's primary objective is to improve traffic efficiency and safety by adjusting the vehicle's speed and position in a predictive manner. Additionally, it considers the dynamic constraints of the traffic environment and the vehicle's operational limitations, aiming to minimize the impact of traffic oscillations on the ego vehicle and, by extension, the surrounding traffic flow. The MPC continuously updates using the latest vehicle state, and together with the upper-layer VLM, this method helps achieve closed-loop state feedback.

### 2.3.1 CAV Dynamic Model

This study focuses on the longitudinal kinematics of vehicles along the driving direction. The key parameters $\theta_t, t \in \{\Delta t^u, 2\Delta t^u, \cdots\}$ obtained from the upper-layer VLM are updated with a time interval of $\Delta t^u$. The lower-layer autonomous driving requires controlling the vehicle at a finer time interval $\Delta t^l$, which is usually around 0.1 second, $\Delta t^l \ll \Delta t^u$. The lower-layer MPC predicts the system's behavior over a finite horizon $N$ and plans the action over this horizon. The output of the MPC at time $t$ is $\{u_t\}_{t \in \{t_0, t_0+\Delta t^l, \cdots, t_0+N \cdot \Delta t^l\}}$. The detailed MPC algorithm is as follows.

The nonlinear longitudinal dynamic model of a single vehicle is described as:

$$\begin{cases} \dfrac{dx_t}{dt} = v_t \\ \dfrac{dv_t}{dt} = a_t \\ \dfrac{da_t}{dt} = f(v_t, a_t) + g(v_t)\eta \end{cases}, t \in \mathcal{T} \quad (9)$$

where $x_n, v_n, a_n$ are position, speed, and acceleration of the vehicle $n$ respectively, and $\eta$ is the engine input. Functions $f$ and $g$ are given by

$$f(v_t, a_t) = -\frac{2K_d}{m} v_t a_t - \frac{1}{\tau^A}\left[a_t + \frac{K_d}{m} v_t^2 + \frac{d_m}{m}\right] \quad (10)$$

$$g(v_t) = \frac{1}{m\tau^A} \quad (11)$$

where $K_d$ represents the aerodynamic drag coefficient, $m$ the vehicle mass, $\tau^A$ is the engine time lag, and $d_m$ is the mechanical drag. In this paper, we focus on the longitudinal kinematics of vehicles. Assuming the parameters in (4) (5) are priori known, we adopt the following control law structure to implement feedback linearization:

$$\eta = mu_t + K_d v_t^2 + d_m + 2\tau^A K_d v_t a_t \quad (12)$$

where $u_t$ is the desired acceleration, determined by the upper controller. Thus, the differential equation of acceleration can be rewritten as:

$$\dot{a}_t = \frac{u_t - a_t}{\tau^A} \quad (13)$$

The objectives of CAV planning are following its preceding vehicle with a desired spacing distance and ensuring safety. Therefore, a constant time headway (CTH) spacing strategy was applied. The desired spacing distance of vehicle $n$ is $d_t^d = h^d v_t + d_0$, $h^d$ and $d_0$ are the desired constant headway and space at a standstill. Based on the CTH rule, the position error $\Delta x_t$ with respect to a desired distance from the preceding vehicle $\Delta x_t = x_t^{-1} - x_t - d_t^d$, where $x_t^{-1}$ is the position of the preceding vehicle/stop line, and $l^{-1}$ is the length of the preceding vehicle.

$$x_t^{-1} = \begin{cases} \min(x_t^{\text{pre}} - l^{-1}, x_t^{\text{SL}}), & \text{if both } x_t^{\text{pre}}, x_t^{\text{SL}} \text{ exit} \\ x_t^{\text{pre}} - l^{-1}, & \text{if only } x_t^{\text{pre}} \text{ exits} \\ x_t^{\text{SL}}, & \text{if only } x_t^{\text{SL}} \text{ exits} \end{cases} \quad (14)$$

If the information about a preceding vehicle or stop line is available, the cost function also includes a term to maintain a safe headway distance:

$$\min_{u[t_0, t_0+N]} J = \sum_{k=0}^{N-1} \left\| \Delta v_{t_0+k \cdot \Delta t^1} \right\|_Q^2 + \sum_{k=0}^{N-1} \left\| u_{t_0+k \cdot \Delta t^1} \right\|_R^2 + \sum_{k=0}^{N-1} \left\| \Delta x_{t_0+k \cdot \Delta t^1} \right\|_{Q^h}^2 + \varepsilon^T \rho \varepsilon \quad (15)$$

where $N$ the predictive horizon length. $Q$, $R$ and $Q^h$ are the weight matrix of error and input, respectively. $\varepsilon$ is the slack vector, and $\rho$ its weight.

When the preceding vehicle or stop line is unavailable, the objective of the MPC is to minimize a cost function $J$ that accounts for the desired speed and control effort over a prediction horizon $N$. The speed error is the gap between vehicle speed and the desired speed $\Delta v_t \coloneqq v_t - v^d$. The cost function is defined as:

$$\min_{u[t_0, t_0+N]} J = \sum_{k=0}^{N-1} \left\| \Delta v_{t_0+k \cdot \Delta t^1} \right\|_Q^2 + \sum_{k=0}^{N-1} \left\| u_{t_0+k \cdot \Delta t^1} \right\|_R^2 + \varepsilon^T \rho \varepsilon \quad (16)$$

The optimization objective is subjected to the following constraints:

$$v_{\min} - \varepsilon \sigma_{\min}^y \le v_t \le v_{\max} + \varepsilon \sigma_{\max}^y \quad (17)$$

$$u_{\min} - \varepsilon \sigma_{\min}^u \le u_t \le u_{\max} + \varepsilon \sigma_{\max}^u \quad (18)$$

## 3 EXPERIMENT

### 3.1 Experiment settings

The experiment settings are given in this section. The proposed methodology is validated using a series of experiments. First, a real-world driving environment dataset is adopted to evaluate the proposed method and baseline methods. We utilize Llava 1.6 (Liu et al., 2024) as the foundation VLM for various components in our system. Additionally, in the sensitivity analysis, we tested the performance of GPT-4-o and GPT-4-mini. Llava 1.6 model was run on a Linux (Ubuntu) system equipped with 32GB of DDR5 RAM and an NVIDIA GeForce RTX 4090 GPU (24GB GDDR6X). The model was configured with a temperature of 0, while all other parameters were left at their default values. The GPT series models were utilized via API calls.

*3.1.1 Dataset*

The nuScenes dataset is a large-scale, real-world autonomous driving dataset containing various locations and weather conditions (Caesar et al., 2020). This dataset includes data from Boston and the US, which follow right-hand traffic rules, and Singapore, which follow left-hand traffic rules. Our proposed algorithm is designed for right-hand traffic. To ensure broader applicability across various environments, we have included some nighttime scenes from Singapore in our evaluation, covering a total of 303 scenes.

The heatmap in Figure 4 visualizes the p-values obtained from analyzing the differences in driving parameters across various scenarios from the nuScenes. The parameter $Q$ does not have results because $Q$ is fixed at 1. Significant differences are indicated with asterisks, suggesting that different parameters should be used when the corresponding features: rain, intersection, and night. Based on these findings, TABLE 2 lists the aggregated parameters selected for eight grouped scenarios. Each scenario is characterized by the presence or absence of rain, intersection, and night conditions. The table provides the final parameters for each of these grouped scenarios.

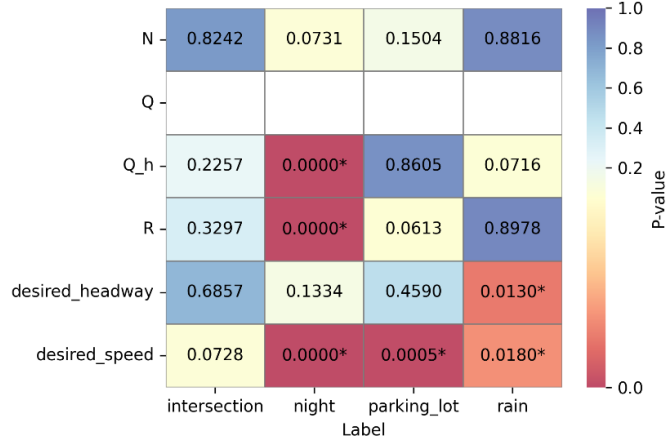

Figure 4 P-values matrix obtained from analyzing the differences in MPC parameters across various scenarios.

TABLE 2 Aggregated parameter of eight grouped scenarios.

| No. | rain | intersection | night | $N$ | $Q$ | $R$ | $Q^h$ | $v^d$ | $h^d$ |
|---|---|---|---|---|---|---|---|---|---|
| 1 | 0 | 0 | 0 | 9 | 1 | 1.68 | 2.75 | 6.44 | 2.60 |
| 2 | 0 | 0 | 1 | 9 | 1 | 1.68 | 1.99 | 5.09 | 2.55 |
| 3 | 0 | 1 | 0 | 9 | 1 | 1.68 | 2.75 | 6.44 | 2.60 |
| 4 | 0 | 1 | 1 | 9 | 1 | 1.15 | 1.99 | 5.09 | 2.55 |
| 5 | 1 | 0 | 0 | 9 | 1 | 1.15 | 1.99 | 5.09 | 2.55 |
| 6 | 1 | 0 | 1 | 9 | 1 | 1.68 | 2.75 | 6.44 | 2.60 |
| 7 | 1 | 1 | 0 | 9 | 1 | 1.68 | 2.75 | 6.44 | 2.60 |
| 8 | 1 | 1 | 1 | 9 | 1 | 1.68 | 1.99 | 5.09 | 2.55 |

*3.1.2  Closed-loop Evaluation Metrics*

We measure three key metrics for closed-loop evaluation from three perspectives: safety, driving comfort, and stability as follows.

1. **Safety: post encroachment time (PET).** PET is a safety metric that measures the time interval between the ego vehicle and another object or vehicle encroaching into the same space (Peesapati et al., 2018). A higher PET value indicates safer driving behavior, reflecting a longer time gap between potential conflicts.
2. **Smoothness: root mean square of acceleration ($RMS^a$).** $RMS^a$ reflects the smoothness of the trajectory. It is calculated as:

$$\text{RMS}^a = \sqrt{\frac{1}{|T|} \sum_{t=0}^{T} a_t^2} \qquad (19)$$

3. **Stability: completion rate.** This metric measures the overall completion rate of scenes, focusing specifically on the ability of the VLM/LLM to generate reasonable and effective outputs consistently. Given that the MPC component is designed with a robust algorithm to avoid infeasible solutions, no MPC failures were observed during the experiments. Therefore, this metric primarily reflects the capability of the VLM/LLM to produce valid and reasonable parameters that enable successful task completion.

*3.1.3 Baseline Models*

To validate the effectiveness of our proposed method, we set up three baseline models. Below are detailed explanations of each baseline model.

**MPC**: This baseline uses the average parameters for the given scenario derived from historical data. The parameters are fixed and do not adapt to the real-time environment. This model does not use any visual information from the vehicle's front camera. Instead, it relies on pre-calculated average parameters without leveraging dynamic memory aggregation. Although it operates in a closed-loop control manner, the control parameters remain static and do not update in real time based on feedback.

**LLM to Action**: This model is based on the approach proposed in (Mao et al., 2024), representing the "FM to Action" methodology in autonomous driving. The method leverages LLMs as cognitive agents to integrate human-like reasoning and decision-making into autonomous driving systems. This enables the LLM to directly generate driving actions.

**LLM-MPC**: This baseline integrates the LLM but does not use the advanced features of our proposed method. It serves as a comparison to assess the basic capabilities of the LLM. Unlike the proposed model, this baseline does not utilize front-camera images for decision-making. Additionally, it does not aggregate parameters from memory based on environmental characteristics, resulting in less adaptive behavior. Like the Baseline MPC Controller, it operates in a closed-loop control manner without real-time parameter updates, as the parameters are initially set and remain unchanged throughout the scenario.

The differences between the proposed VLM-MPC and the two baseline models are outlined in TABLE 3. By comparing these models, we aim to demonstrate that our proposed LLM Controller, which utilizes image inputs and reference memory, offers significant advantages in navigating complex driving scenarios and making more informed and adaptive decisions. In the future, we will conduct additional ablation studies to validate the importance of image input, memory, and reflection independently.

TABLE 3 Comparison of Proposed VLM-MPC and two baseline models

| Controllers | Image input | Memory | Reflection |
|---|---|---|---|
| MPC | No | No | Closed-loop |
| LLM to Action | No | Yes | Open-loop |
| LLM-MPC | No | Yes | Closed-loop |
| VLM-MPC | Yes | Yes | Closed-loop |

## 3.2 Experiment Results

Based on the experiment design, this section comprehensively evaluates the proposed VLM-MPC, focusing on safety, smoothness, and completion rate across various driving scenarios.

*3.2.1 Safety*

The safety evaluation results using PET as metrics are shown in TABLE 4. The proposed VLM-MPC demonstrates superior performance compared to the baseline MPC controller, baseline LLM controller, and real-world trajectory regarding minimum PET values across various traffic conditions. Notably, the VLM-MPC consistently maintains PET values above the critical safety threshold of 1 second in all scenarios, which signifies better safety margins. When comparing the results, it is evident that the proposed VLM-MPC exhibits greater stability and adaptability in both dry and rainy conditions and at intersections and parking lots. For instance, in the absence of rain and at non-intersections, the VLM-MPC achieves PET values of 1.31 and 1.97, respectively, which are higher than those of the real-world trajectory (1.65 and 1.33), baseline MPC controller (1.73 and 1.74), and baseline LLM controller (0.37 and 0.05). In more challenging conditions such as rainy intersections, the VLM-MPC still performs robustly with PET values of 1.36 and 1.92, outperforming the real-world trajectory (4.83), baseline MPC controller (5.21), and baseline LLM controller (0.64). The consistent performance across all tested conditions underscores the VLM-MPC's ability to enhance vehicle safety by maintaining PET values above the critical threshold, thereby substantially improving existing methods.

TABLE 4 Minimum PET (s) for the grouped situation of proposed VLM-MPC and baseline models.

| | Scenario Number | 1 | 2 | 3 | 4 | 5 | 6 | 7 | 8 |
|---|---|---|---|---|---|---|---|---|---|
| Scenario property | Rain | 0 | 0 | 0 | 0 | 1 | 1 | 1 | 1 |
| | Night | 0 | 0 | 1 | 1 | 0 | 0 | 1 | 1 |
| | Intersection | 0 | 1 | 0 | 1 | 0 | 1 | 0 | 1 |
| Minimum PET (s) of different controller | Real-world | 1.65 | 1.53 | 1.33 | 4.61 | 3.15 | 2.6 | 2.85 | 3.56 |
| | MPC | 1.12 | 1.1 | 1.33 | 1.26 | 1.91 | 1.62 | 1.65 | 1.2 |
| | LLM to Action | 0.45 | 1.59 | 1.26 | 1.36 | 1.36 | 2.65 | 0.75 | 1.84 |
| | LLM-MPC | 1.11 | 1.21 | 1.34 | 1.07 | 2.13 | 1.19 | 3.13 | 1.10 |
| | VLM-MPC | 1.31 | 1.18 | 1.97 | 1.03 | 2.68 | 1.36 | 3.53 | 1.92 |

Figure 5 demonstrates the application of the proposed VLM-MPC in a rainy intersection scenario in Boston. The Figure highlights the dynamic and adaptive nature of the VLM-MPC as it navigates the vehicle through the environment.

At $t = 0s$, the VLM agent receives the initial scene input, including basic instruction, reference memory, vehicle front camera image, environment description, and ego/preceding vehicle state. Based on this input, the VLM agent generates the initial key driving parameters [10,1.0,2.0,3.5,6.5,2.8]. These parameters are optimized to maintain safety and efficiency, considering the foggy weather and the presence of parked vehicles.

At $t = 5s$, the updated scenario shows the vehicle approaching an intersection with the preceding vehicle no longer detected. The ego vehicle's position and speed have changed, indicating it is getting closer to the intersection. The VLM agent recognizes this proximity to the intersection and updates the key driving parameters to [12,1,1.0,0.5,2.0]. The critical adjustment here is the reduction in the desired speed (from 6.5 m/s to 2.0 m/s). This change is crucial as it ensures the vehicle gradually slows down while approaching the intersection, enhancing safety by allowing it to stop smoothly if necessary.

By $t = 10s$, the vehicle is even closer to the intersection, and the VLM agent continuously updates the parameters to ensure the vehicle decelerates appropriately. The model's ability to adjust the desired speed based on real-time feedback demonstrates its effectiveness in handling complex driving scenarios. This adaptive behavior is essential for navigating intersections safely, especially under challenging weather conditions like rain. Due to space limitations, the vehicle's procedures at t=10 and t=15 are not shown in Figure 3.

The example case in Figure 5 illustrates the VLM-MPC system's capability to dynamically adjust key driving parameters in response to changing environmental and situational contexts, ensuring safe and efficient vehicle control. The system's proactive reduction in desired speed as the vehicle approaches the intersection at $t = 5s$ exemplifies its intelligent and adaptive control strategy. If the vehicle keeps applying the driving parameters given by GPT at $t = 5s$, the large desired speed, it will not stop safely and smoothly before the stop line.

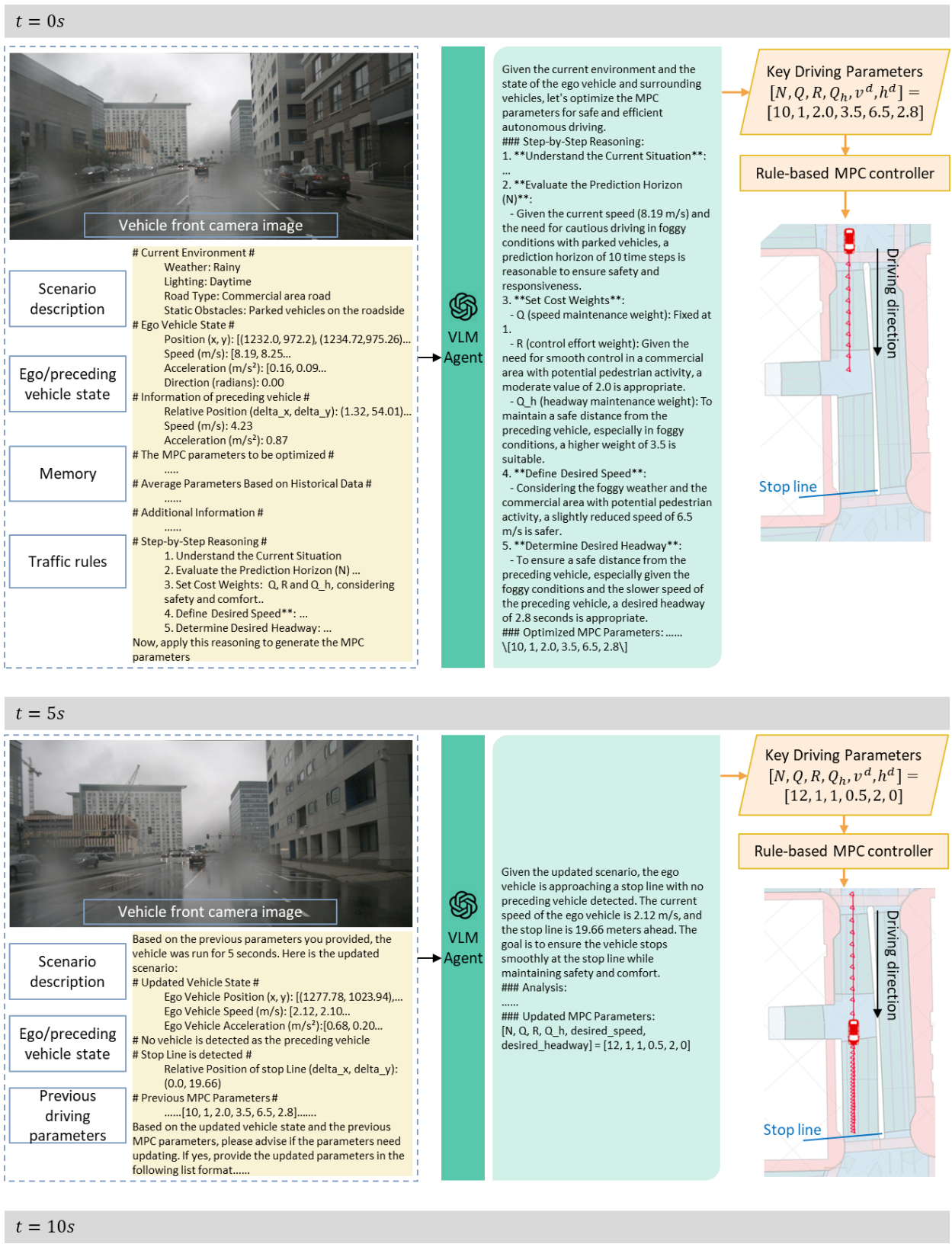

Figure 5 Illustration of the proposed VLM-MPC at one rainy intersection scenario in Boston.

Figure 6 illustrates two unsafe scenarios generated by the Baseline LLM to Action Controller. The trajectory shows snapshots from the final 3 seconds before a collision. In Figure 6 (a), during the daytime, the ego vehicle is approaching a situation where two vehicles are moving in the left lane ahead, while on the right, there are two parked vehicles and a work zone. In this case, the Baseline LLM to Action Controller plans a trajectory that advances forward, which is clearly unreasonable as it risks a collision with the parked cars. In contrast, the proposed VLM-MPC slows down to avoid potential collision. In Figure 6 (b), during nighttime, a similar issue arises: the LLM to Action Controller fails to account for a vehicle stopped at an intersection and instead accelerates towards it, resulting in a PET of 0.75 seconds, which is below the safety threshold. The proposed VLM-MPC, on the other hand, appropriately reduces speed to avoid collision.

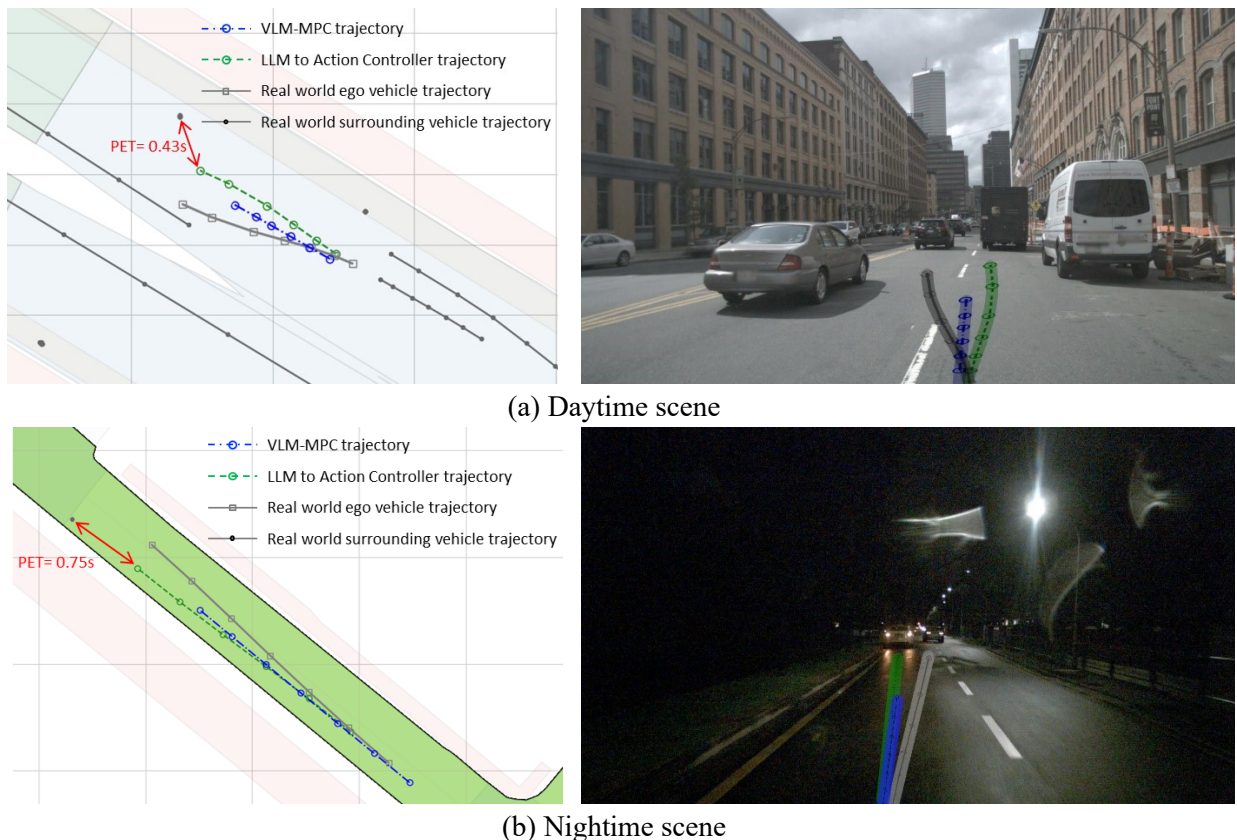

(a) Daytime scene

(b) Nightime scene

Figure 6 Comparison of the vehicle trajectory using proposed VLM-MPC and baseline LLM to Action controller

### 3.2.2 Smoothness

TABLE 5 illustrates the driving comfort of the proposed VLM-MPC system compared to baseline models by showing the $RMS^a$ across various scenarios. The proposed VLM-MPC system consistently achieves the average lowest $RMS^a$ values across most scenarios compared to the baseline models. This indicates that the VLM-MPC system provides smoother and more comfortable driving experiences. The reduction in $RMS^a$ suggests that the VLM-MPC effectively minimizes abrupt changes in acceleration, leading to a more stable and pleasant ride for passengers.

The Baseline LLM to Action Controller exhibits significantly highest $RMS^a$ comparing to real-world trajectory and other models, due to the generated trajectories not aligning with vehicle dynamics. Figure 7 shows the distribution of speed and acceleration across different methods. It is evident that all the proposed VLM-MPC, the baseline MPC, and the baseline LLM-MPC, maintain vehicle acceleration within the range of -3.5 to 3 m/s², which is achieved through lower-level MPC optimization, and their speed and acceleration distributions closely match the real-world trajectory. The real-world trajectory includes an extreme deceleration of -10.53 m/s² due to vehicle positioning errors. The Baseline LLM to Action Controller generates a trajectory with an extreme deceleration of -23.2 m/s². This highlights the inability of the LLM to directly plan a trajectory that adheres to vehicle kinematics and dynamics, making it difficult for the vehicle to execute the planned actions effectively.

TABLE 5 Smoothness RMS$^a$ for the grouped situation of proposed VLM-MPC and baseline models.

| | Scenario Number | 1 | 2 | 3 | 4 | 5 | 6 | 7 | 8 |
|---|---|---|---|---|---|---|---|---|---|
| Scenarios | Rain | 0 | 0 | 0 | 0 | 1 | 1 | 1 | 1 |
| | Night | 0 | 0 | 1 | 1 | 0 | 0 | 1 | 1 |
| | Intersection | 0 | 1 | 0 | 1 | 0 | 1 | 0 | 1 |
| Smoothness RMS$^a$ of different controller | Real-world | 0.61 | 0.54 | 0.63 | 0.56 | 0.56 | 0.51 | 0.68 | 0.58 |
| | MPC | 0.41 | 0.44 | 0.47 | 0.51 | 0.4 | 0.42 | 0.35 | 0.46 |
| | LLM to Action | 0.43 | 0.93 | 0.94 | 1.41 | 2.14 | 2.18 | 3.13 | 1.66 |
| | LLM-MPC | 0.35 | 0.4 | 0.48 | 0.5 | 0.38 | 0.4 | **0.33** | 0.48 |
| | VLM-MPC | **0.33** | **0.4** | **0.42** | **0.43** | **0.38** | **0.39** | 0.34 | **0.4** |

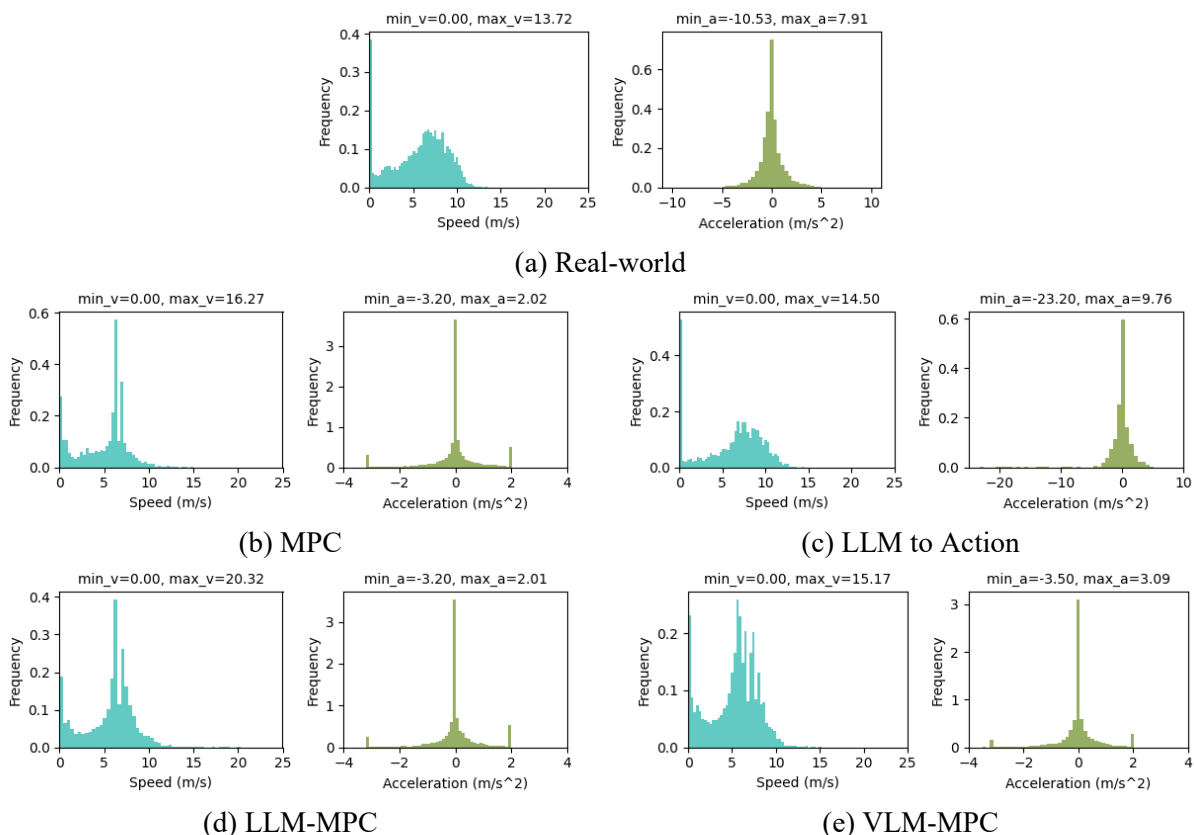

(a) Real-world

(b) MPC

(c) LLM to Action

(d) LLM-MPC

(e) VLM-MPC

Figure 7 Distribution of speed and acceleration of proposed VLM-MPC and baseline models.

### 3.2.3 Completion Rate

VLM typically suffers from arbitrary predictions, which can produce invalid outputs (e.g., hallucinations or invalid formats) that are detrimental to driving systems. TABLE 6 listed the completion rate of proposed VLM-MPC using various FM Models and baseline models. The completion rate for Llava 1.6 across all scenarios was 99.7%, while the GPT series models (GPT-3.5-turbo, GPT-4-o, GPT-4-mini) achieved a 100% completion rate. This difference can be attributed to the significantly larger number of parameters in the GPT series models compared to Llava 1.6. Llava 1.6 uses a 7B parameter model, whereas GPT-3.5-turbo has 175B parameters. The GPT-4-o model is estimated to have between 100B and 1T parameters, and GPT-4-mini is estimated to have between 10B and 30B parameters. The larger parameter size in the GPT models enables them to handle a wider range of scenarios and produce more reliable outputs, reducing the likelihood of errors such as hallucinations or invalid format outputs. This robustness in handling diverse tasks contributes to the 100% completion rate observed in the GPT series models.

TABLE 6 Completion rate of different FM models

| Models | FM Model | FM Model Parameter Size | Completion rate (%) |
| --- | --- | --- | --- |
| LLM-MPC | Llama-3.1 | 7B | 99.7 |
|  | GPT-3.5-turbo | 175B | 100 |
| LLM to Action | GPT-3.5-turbo | 175B | 100 |
| VLM-MPC | Llava-1.6 | 7B | 99.7 |
|  | GPT-4-o | 100B~1T | 100 |
|  | GPT-4-mini | 10B~30B | 100 |

### 3.2.4 Response Time

Figure 8 illustrates the response speeds of the proposed VLM-MPC with different models. The Llava model has an average response time of 3.42 seconds, meeting the system's requirement of a 0.2Hz response speed. In contrast, the GPT-4-o and GPT-4-mini models have average response times of around 10 seconds, primarily because their parameter sizes are several orders of magnitude larger than Llava model.

Figure 9 shows the response speeds of the baseline LLM-MPC with different models. Llama 3.1 and GPT-3.5-turbo models were tested. These models do not require image input, so the average response time is shorter than that of vision input models. The Llama 3.1 model has an average response time of 1.5 seconds, which is significantly faster than GPT-3.5-turbo's average response time of 3.9 seconds. This difference is also due to the smaller parameter size of Llama 3.1. The smaller parameter size allows for quicker processing and response times, making Llama 3.1 more suitable for applications where rapid decision-making is crucial. Conversely, while GPT-3.5-turbo has a larger parameter size that can handle more complex scenarios, it requires more computational resources, resulting in longer response times.

Overall, the response speed of the two locally running models (Llava and Llama) is faster than that of the cloud-based GPT models. This is primarily due to the difference in model parameter sizes, and also because the GPT models are affected by network latency and account token limits. In the future, we plan to fine-tune smaller parameter models to prepare for their application in vehicles.

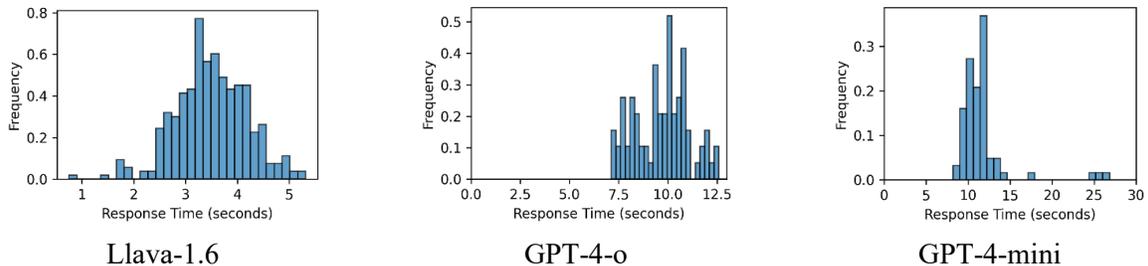

Llava-1.6      GPT-4-o      GPT-4-mini

Figure 8 Distribution of response times of proposed VLM-MPC

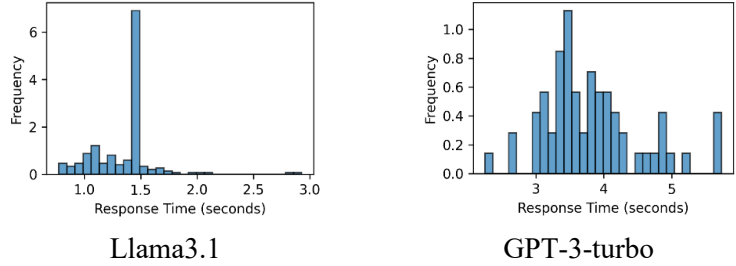

Llama3.1            GPT-3-turbo

Figure 9 Distribution of response times of baseline LLM-MPC

### 3.3 Sensitivity analysis

To validate the effectiveness of the proposed model, this section examines the impact of its key components on performance. And the sensitivity analysis on response time.

#### 3.3.1 Impact of Reference Memory

The ablation model from the proposed VLM-MPC, which excludes the reference memory as VLM input (VLM-MPC without M), achieves a completion rate of 99.3%. This model experienced two instances of invalid responses, as detailed in TABLE 7. The response time remains around 1.5 seconds. In each scene, safety was maintained, as shown in TABLE 8. However, the parameters generated by the VLM generally favored higher speeds and larger headways. While the planned trajectories led to improved smoothness under clear weather conditions, the smoothness deteriorated in rainy and nighttime scenarios.

TABLE 7 Completion rate when using Llava-1.6.

|  | Completion rate (%) |
|---|---|
| VLM-MPC without E | 87.5 |
| VLM-MPC without M | 99.3 |
| VLM-MPC | 99.7 |

TABLE 8 Smoothness RMS$^a$ for the grouped situation of proposed VLM-MPC and baseline models

| Scenario Number | | 1 | 2 | 3 | 4 | 5 | 6 | 7 | 8 |
|---|---|---|---|---|---|---|---|---|---|
| Scenarios | Rain | 0 | 0 | 0 | 0 | 1 | 1 | 1 | 1 |
| | Night | 0 | 0 | 1 | 1 | 0 | 0 | 1 | 1 |
| | Intersection | 0 | 1 | 0 | 1 | 0 | 1 | 0 | 1 |
| Minimum PET (s) of different controller | VLM-MPC without M | 2.53 | 1.36 | 1.86 | 1.39 | 2.14 | 2.32 | 14.95 | 1.20 |
| | VLM-MPC without E | 1.66 | 1.13 | 1.58 | 3.70 | 2.28 | 1.91 | 1.65 | 1.20 |
| | VLM-MPC | 1.31 | 1.18 | 1.97 | 1.03 | 2.68 | 1.36 | 3.53 | 1.92 |
| Smoothness RMS$^a$ of different controller | VLM-MPC without M | 0.35 | 0.40 | 0.41 | 0.46 | 0.35 | 0.30 | 0.74 | 0.95 |
| | VLM-MPC without E | 0.37 | 0.40 | 0.47 | 0.35 | 0.36 | 0.36 | 0.77 | 0.91 |
| | VLM-MPC | 0.33 | 0.4 | 0.42 | 0.43 | 0.38 | 0.39 | 0.33 | 0.4 |

#### 3.3.2 Impact of Environment Description

The ablation model from the proposed VLM-MPC, which excludes the environment encoder as VLM input, is called 'VLM-MPC without E'. When environmental information is excluded from the prompt to VLM, the completion rate drops significantly to 87.5%, much lower than the proposed VLM-MPC method, shown in TABLE 7. The observed failures primarily occur because without road information, the VLM fails to comprehend the current environment and driving task, resulting in instances where no parameters are output, or the output parameters do not match the specified format. This suggests that road information plays a critical role in completing the description of the driving task, providing the VLM with comprehensive context needed to understand and accurately respond to the driving task.

In terms of response time, excluding environment information noticeably slows down Llama's response speed, with some instances exceeding 200 seconds. This excessively long response time further underscores the importance of environment descriptions in helping the VLM understand scene contexts.

In the scenarios where responses were successful, each scene could maintain safety, and compared to real-world trajectories and end-to-end methods, the response time was still significantly shorter.

*3.3.3 Interpretability analysis*

To facilitate a more intuitive and comprehensible interpretation of the VLM-MPC, aiding in the analysis of the causal relationships between the environment and VLM outputs. We demonstrate this with two selected scenarios.

First, we extracted the model's attention allocation to the image input while generating each token. A higher sum of attention weights indicates a greater focus on the image content when generating that particular token (zjysteven, 2024). Figure 10 presents the sum of attention the model assigns to the image input during the generation of each token, with the horizontal axis representing the tokens in sequence. The results reveal a consistent pattern across the two scenarios: during the first 50-60 tokens, approximately 14% of the attention is directed towards the vision tokens, significantly higher than in subsequent tokens. This is aligned with our implementation of CoT, where the initial step is 'Understand the current environment'. Consequently, the model's initial 50-60 tokens are dedicated to describing the environment, naturally leading to a stronger correlation with the image input. The subsequent tokens correspond to steps two through five of the CoT: 'Evaluate the prediction horizon', 'Set cost weights', 'Define desired speed', and 'Determine desired headway'. These steps are more inference-focused and rely less on the image input, drawing more from textual information.

Subsequently, we further integrated the vision encoder to visualize the attention map over the input image. The vision encoder in the model processes the image input and generates attention weights across different layers. By averaging the attention weights across all layers, we obtained a comprehensive attention matrix. For each key phrase generated, we combined the model's attention weights with the vision encoder's attention matrix to compute the model's focus on different regions of the image during the generation of that phrase. The computed attention values were then mapped onto the input image (vehicle front camera image), highlighting the specific regions the model attended to in the image.

Figure 10 illustrates the results for two representative scenarios, where we selected the first few tokens generated by the model and the corresponding short phrases, along with the attention allocated to the image during their generation. In the first scenario, when the model outputs "… with vehicles driving on a road…", the attention is concentrated on the road ahead. When the model generates "…there are parked vehicles…", more attention is directed towards the parked vehicles on the side of the road. When it outputs "…The road appears to be wet…", the attention becomes more dispersed, with less focus on the vehicles. In the second scenario, when the model generates "…a traffic light is visible…", the traffic light in the image clearly receives the highest attention. This demonstrates that the model indeed focuses on relevant parts of the image when generating these descriptions, highlighting the importance of the visual input.

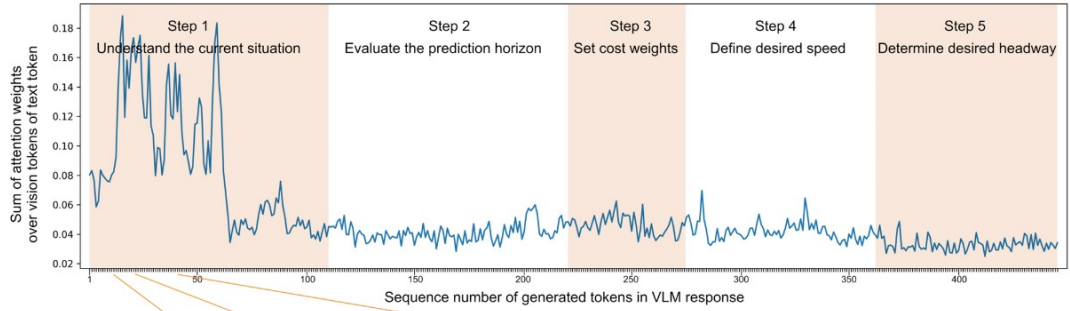

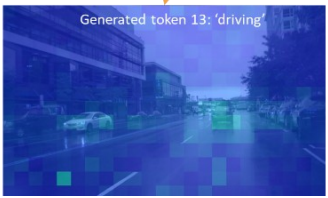 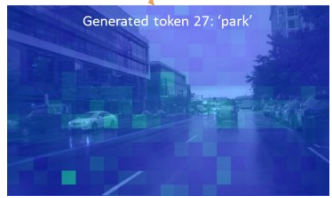 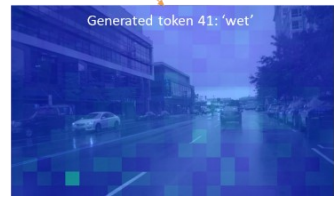

(a) Scene 1

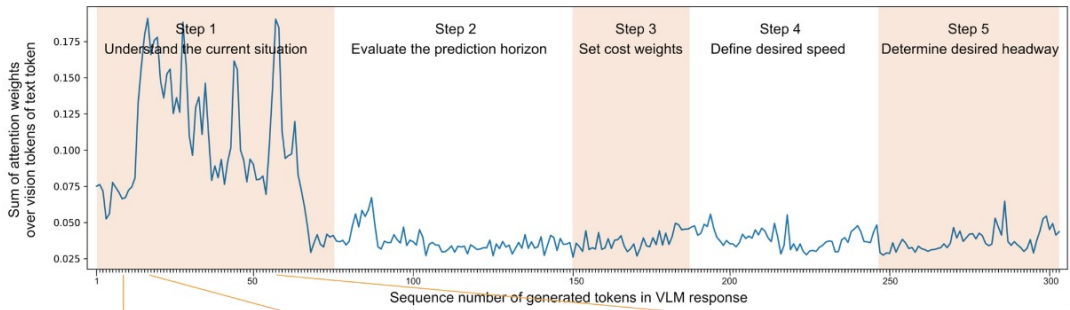

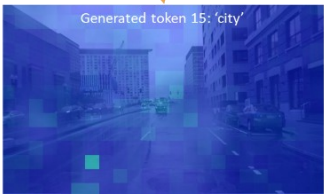 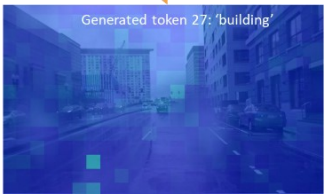 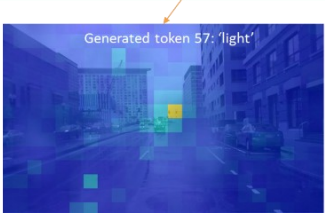

(b) Scene 2

Figure 10 Sum of attention the model pays to the image input when generating each token.

## 4 CONCLUSION AND FUTURE RESEARCH

This paper introduced a closed-loop autonomous driving controller, VLM-MPC, that leverages VLMs for upper-level decisions and MPC for low-level vehicle control. This design enables the upper-level model to provide adaptive decision-making based on scenario changes, while the lower level controls the vehicle in real-time based on the current state of the ego vehicle and surrounding vehicles.

In experiments based on the nuScenes dataset, we compared this structure with the method of using an LLM to control the vehicle independently. The results validate that a model-based approach like MPC significantly enhances the performance of LLM-based controllers. The integration of MPC allows the system to make more accurate and reliable decisions, especially in dynamic and complex driving environments, demonstrating the advantages of combining model-based methods with LLMs for autonomous driving.

Experiment results demonstrate that the VLM-MPC consistently maintains PET above safe thresholds, in contrast to some scenarios where the VLM-based control posed collision risks. Additionally, the VLM-MPC enhances smoothness compared to the real-world trajectories and VLM-based control. By comparing behaviors under different environmental settings, we highlight the VLM-MPC's capability to understand the environment and make reasoned inferences. VLM-MPC with Llava could reach the response time requirements of the proposed method. Moreover, we validate the significant contributions of two key components, the reference memory and the environment encoder, to the stability of responses through ablation tests.

Future research will focus on two main directions. The first direction is to refine the proposed approach with more ablation studies to validate the effectiveness of each component and further optimize the system. This includes investigating the relationship between the VLM's image input understanding capabilities and the subsequent driving behavior choices, assessing the impact of memory on the stability of the VLM-generated results, and evaluating how the frequency of the asynchronous operation between the upper and lower layers affects overall vehicle performance. The second direction is to conduct real-world vehicle experiments. These experiments will help collect data on specific roads under various driving conditions. While datasets like nuScenes and Waymo provide extensive information on real-world scenarios, they often lack data on the same scene under different driving conditions. Collecting our own data will enable us to better analyze the impact of driving environments on decision-making. This will also help identify out-of-distribution (ODD) scenarios involving adverse weather conditions or low-light environments. Additionally, real-world experiments offer the most rigorous closed-loop testing, considering the cumulative uncertainties from perception, decision-making, and low-level vehicle control systems.

## 5 ACKNOWLEDGMENT

This work was supported by the National Science Foundation Cyber-Physical Systems (CPS) program. Award Number: 2343167.